\documentclass[10pt,twocolumn,letterpaper]{article}

\usepackage{titling}

\usepackage[pagenumbers]{cvpr} 

\usepackage{graphicx}
\graphicspath{{assets/}} 
\usepackage{amsmath}
\usepackage{amssymb}
\usepackage{booktabs}

\usepackage{fancyref}
\usepackage{cite}
\DeclareMathAlphabet\mathcalbf{OMS}{cmsy}{b}{n}
\usepackage{mathabx}
\usepackage{mathtools}
\usepackage[utf8]{inputenc}
\usepackage[T1]{fontenc}
\usepackage{float}

\usepackage{algorithm}
\usepackage{algpseudocode}
\usepackage{bm}
\usepackage{multirow}
\usepackage[hang,flushmargin]{footmisc}
\usepackage{algorithm}
\usepackage{algpseudocode}
\usepackage{mathtools,nccmath}
\usepackage{cite}
\usepackage{afterpage}

\usepackage[pagebackref,breaklinks,colorlinks]{hyperref}

\usepackage[capitalize]{cleveref}
\crefname{section}{Sec.}{Secs.}
\Crefname{section}{Section}{Sections}
\Crefname{table}{Table}{Tables}
\crefname{table}{Tab.}{Tabs.}

\def\cvprPaperID{} 
\def\confName{CVPR}
\def\confYear{2022}

\begin{document}

\title{Intelli-Paint: Towards Developing Human-like Painting Agents}

\author{Jaskirat Singh$^{1,2}$, Cameron Smith$^{1}$, Jose Echevarria$^1$, Liang Zheng$^2$,\\
$^1$Adobe Research, $^2$Australian National University\\
{\tt\small \{jaskirat.singh,liang.zheng\}@anu.edu.au, \{casmith, 
echevarr\}@adobe.com}
}


\twocolumn[{
\maketitle
\begin{center}
     \vskip -0.1in
    \captionsetup{type=figure}
    \includegraphics[width=1\linewidth]{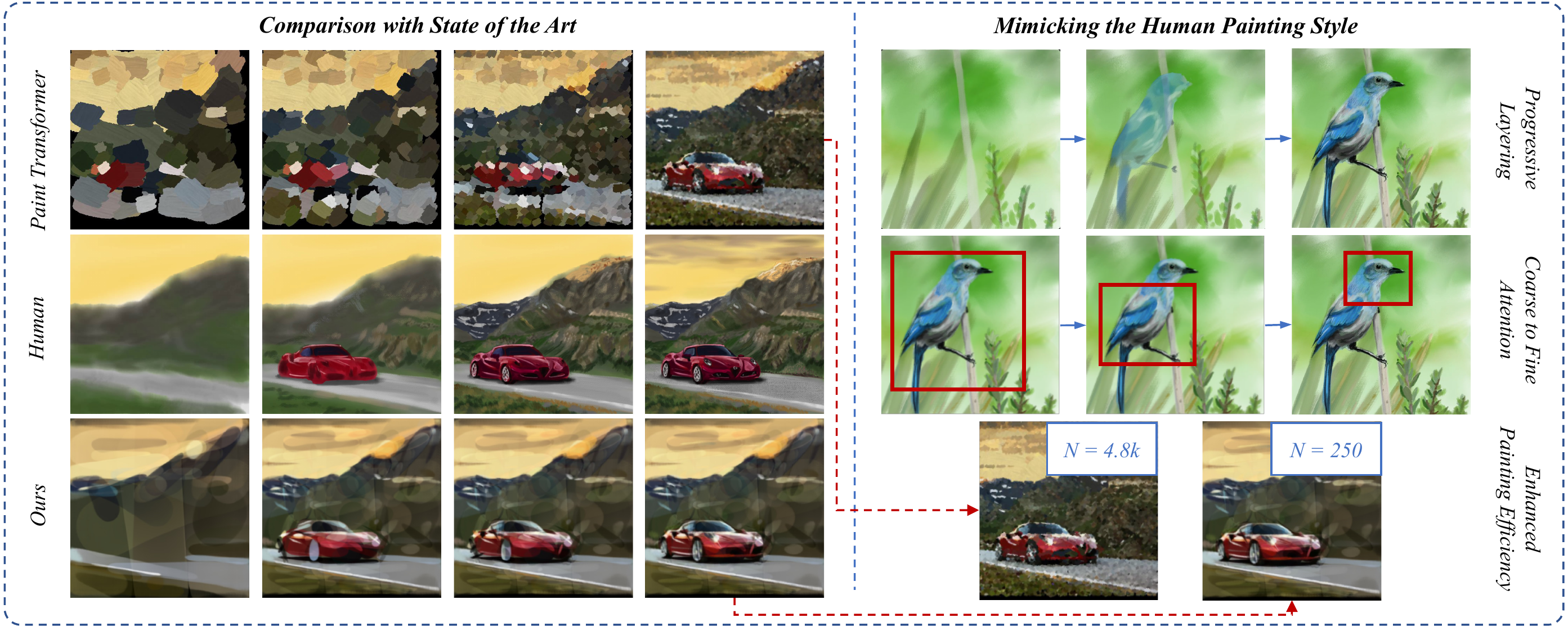}
      \vskip -0.1in
    \captionof{figure}{\textbf{Developing human-like painting style.} \emph{(Left)} Painting sequence visualization which demonstrates that our method exhibits significantly higher resemblance with the human painting style as opposed to previous state of the art. \emph{(Right)} This resemblance is achieved through 1) a progressive layering strategy which allows for a more human-like evolution of the canvas, 2) a sequential attention mechanism which focuses on different image regions in a coarse-to-fine fashion and 3) a brushstroke regularization formulation which allows our method to obtain detailed results while using significantly fewer brushstrokes ($\sim$1/20 as compared to Paint Transformer \cite{liu2021paint} in above).}
     \label{fig:overview}
\end{center}
}]

\begin{abstract}
\vskip -0.1in
   The generation of well-designed artwork is often quite time-consuming and assumes a high degree of proficiency on part of the human painter. In order to facilitate the human painting process, 
   substantial research efforts have been made on teaching machines how to ``paint like a human'', and then using the trained agent as a painting assistant tool for human users. However, current research in this direction is often reliant on a progressive grid-based division strategy wherein the agent divides the overall image into successively finer grids, and then proceeds to paint each of them in parallel. This inevitably leads to artificial painting sequences which are not easily intelligible to human users.
    To address this, we propose a novel painting approach which learns to generate output canvases while exhibiting a more human-like painting style. The proposed painting pipeline Intelli-Paint consists of 1) a progressive layering strategy which allows the agent to first paint a natural background scene representation before adding in each of the foreground objects in a progressive fashion. 2) We also introduce a novel sequential brushstroke guidance strategy which helps the painting agent to shift its attention between different image regions in a semantic-aware manner. 3) Finally, we propose a brushstroke regularization strategy which allows for $\sim$60-80\% reduction in the total number of required brushstrokes without any perceivable differences in the quality of the generated canvases. Through both quantitative and qualitative results, we show that the resulting agents not only show enhanced efficiency in output canvas generation but also exhibit a more natural-looking painting style which would better assist human users express their ideas through digital artwork.
\end{abstract}

\section{Introduction}
\label{sec:intro}

Paintings form a key medium through which humans express their ideas and emotions. Nevertheless, the ability with which such ideas can be expressed is often limited by the prowess of the human painter. Moreover, the creation of finer-quality art is often quite challenging and requires a considerable amount of time on part of the human painter.

One of the ways to address this problem is to develop autonomous painting agents which can assist human painters to better express their ideas in a quick and concise fashion. To this end, there is a growing research interest \cite{huang2019learning,mellor2019unsupervised,ganin2018synthesizing,zheng2018strokenet,kotovenko2021rethinking,xie2013artist,ha2017neural,zou2021stylized,liu2021paint,singh2021combining,nakano2019neural,frans2018unsupervised,frans2021clipdraw} in teaching machines \emph{``how to paint''}, in a manner similar to a human painter. For instance, Huang \etal \cite{huang2019learning} 
use deep reinforcement learning in order to learn an unsupervised brushstroke decomposition for the creation of non-photorealistic imagery. Zou \etal \cite{zou2021stylized} use gradient descent in order to optimize over the brushstrokes parameters for the entire painting trajectory. Similarly, Liu \etal \cite{liu2021paint} propose a novel Paint Transformer which formulates the learning to paint problem as a feed-forward set prediction problem.
Despite their efficacy, the current works often lack semantic understanding of the image contents and are invariably reliant on a progressive grid-based division strategy, wherein the painting agent divides the overall image into successively finer grids, and then proceeds to paint each of them in parallel. This inevitably leads to hierarchically bottom-up painting sequences which are quite mechanical and thus not applicable for human users.

In this paper, we propose a novel painting pipeline \emph{Intelli-Paint} which tries to address the need for semantic-aware painting sequences, by mimicking the human painting style 
in three main ways. 
\textbf{First,} we propose a progressive layering strategy which, much like a human, allows the painting agent to successively draw a given scene in multiple layers. That is, instead of starting to paint the entire scene at once, our method learns to first paint a realistic background scene representation before adding in each of the foreground objects in a progressive layerwise fashion.

\textbf{Second,} the human painting process is characterized by a localized spatial attention span. For instance, a potential artist would focus on different local image areas while painting distinct parts of the final canvas \cite{zhao2020painting}. This is in sharp contrast with previous works, which either focus on the entire image or several predefined grid blocks \cite{liu2021paint,zou2021stylized}. To better mimic the human style, we introduce a sequential brushstroke guidance approach which allows the painting agent to shift its attention between different image areas through a sequence of localized attention windows.
The spatial dimensions and position of the localized attention window are progressively adjusted during the painting process so as to paint a given scene in a coarse-to-fine fashion.

\textbf{Third,} we note that prior works often use a fixed brushstroke budget which not only leads to wasteful / overlapping brushstroke patterns (refer Fig. \ref{fig:human_sim_qual}) but also imparts an artificial painting style to the final agent. 
To this end, we propose an inference-time brushstroke regularization formulation which removes brushstroke redundancies by regularizing the total number of brushstrokes required for painting a given canvas. Our experiments reveal that this not only leads to a $\sim$60-80\% enhancement in the brushstroke decomposition efficiency but also leads to more natural looking painting sequences which are easily intelligible by a human painter.
In summary, towards developing human-like painting agents, this paper makes the following contributions.

\begin{itemize}
    \item We introduce a progressive layering approach, which much like a human, allows the painting agent to draw a given scene in multiple successive layers.
    \item We propose a sequential brushstroke guidance strategy which enables the painting agent to focus on different image regions through a learned sequence of coarse-to-fine localized attention windows.
    \item Finally, we introduce an inference time brushstroke regularization procedure which results in a $\sim$60-80\% enhancement in the brushstroke decomposition efficiency and leads to more natural painting sequences which are easily intelligible by a human user. 
\end{itemize}

\section{Related Work}
\label{sec:related_work}

\begin{figure*}[h!]
\begin{center}
\centerline{\includegraphics[width=\linewidth]{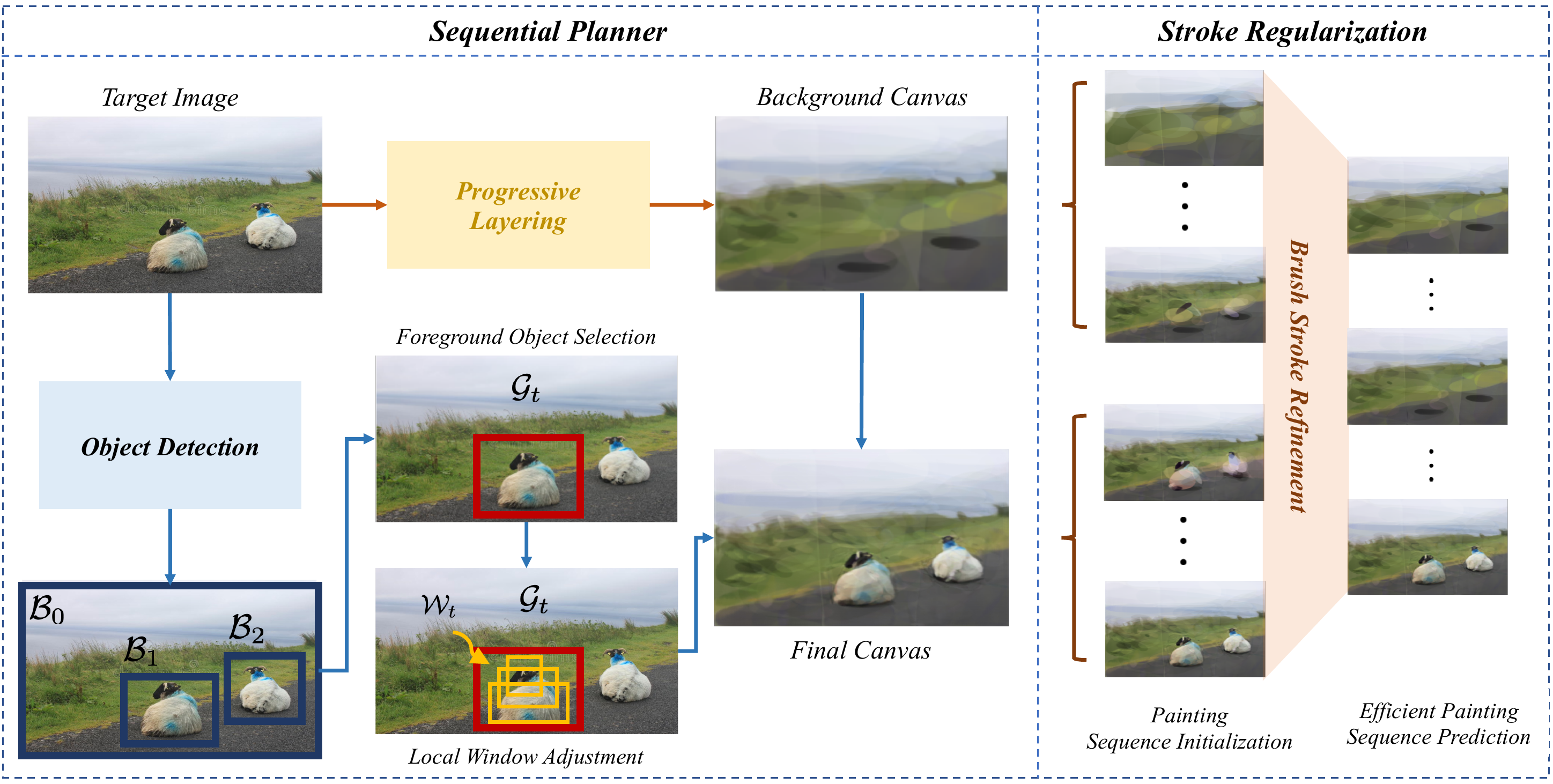}}
\caption{\emph{\textbf{Method Overview.}} Given a target image $I$, the \emph{Intelli-Paint} agent first learns to paint a realistic background scene on the canvas. Once the background scene has been painted, the agent then proceeds to progressively add each of the foreground objects using a sequential brushstroke guidance procedure. To do this, the painting agent first uses the convex combination formulation from Eq.~\ref{eq:fg_obj_sel_2} to select the foreground object it would like to paint (indicated by object window $\mathcal{G}_t$). The features within each object region are then painted in a coarse-to-fine fashion through a sequence of localized attention windows $\mathcal{W}_t$. Finally, the brushstroke sequence is fed into a regularization procedure which removes brushstroke redundancies / overlaps to output the most efficient painting sequence for each test image.}
\label{fig:model_design}
\end{center}
\vskip -0.3in
\end{figure*}

The problem of teaching machines \emph{``how to paint''} has been extensively studied in the context of stroke-based rendering (SBR), which focuses on the recreation of non-photorealistic imagery through appropriate positioning and selection of discrete elements such as paint strokes or stipples \cite{hertzmann2003survey,zeng2009image}. Traditional SBR algorithms often require greedy search \cite{hertzmann1998painterly,litwinowicz1997processing}, optimization over devised heuristics \cite{turk1996image}, or user input for selecting the position and nature of each brushstroke \cite{haeberli1990paint,teece19983d}. More recent solutions \cite{ha2017neural,graves2013generating} adopt the use of recurrent neural networks for computing optimal brushstroke decomposition. 
However, these methods require access to dense human brushstroke annotations, which limits their applicability to most real world problems. In another work, Zhao \etal \cite{zhao2020painting} use a conditional variational autoencoder framework for synthesising time-lapse videos depicting the recreation of a given target image. However, this requires access to painting time-lapse videos from real artists for training. Furthermore, the time-lapse outputs are generated at very low-resolution as compared to the high-resolution output sequences generated using our approach. 

In recent years, there has been an increased focus on learning an unsupervised brushstroke decomposition without requiring access to dense human brushstroke annotations. For instance, recent works \cite{xie2013artist,jia2019lpaintb,ganin2018synthesizing,mellor2019unsupervised,huang2019learning, singh2021combining} use deep reinforcement learning and an adversarial training approach for learning an efficient brushstroke decomposition. Optimization-based methods \cite{zou2021stylized} directly search for the optimal brushstroke parameters by performing gradient descent over a novel optimal-transport-based loss function. In another recent work, Liu \etal \cite{liu2021paint} propose a novel Paint Transformer which formulates the learning to paint problem as a feed-forward stroke set prediction problem. 

While the above works show high proficiency in painting quality output canvases, the generation of the same invariably depends on a progressive grid-based division strategy. In this setting, the agent divides the overall image into successively finer grids, and then proceeds to paint each of them in parallel. Experimental analysis reveals that this not only reduces the efficiency of the final agent, but also leads to mechanical (grid-based) painting sequences which are not easily intelligible to human users.

\section{Our Method}
\label{sec:our_method}

The Intelli-Paint framework (refer Fig.~\ref{fig:model_design}) is based on a two-stage hybrid optimization strategy which consists of two modules: \emph{Sequential Planner} and \emph{Stroke Regularizer}. In the first stage, the \emph{Sequential Planner} learns to predict a coarse but human-like initialization for the brushstroke sequence ${\mathbf{s}}_{init}$. The coarse brushstroke sequence initializations are then fed into a gradient descent based \emph{stroke regularization} procedure,
 which removes redundant brushstroke patterns and refines the original brushstroke parameters to output the most efficient stroke decomposition ${\mathbf{s}}_{pred}$ for each test image. This two-stage process is formulated as,
\begin{align}
    {\mathbf{s}}_{init} = Planner(C_{init},I_{target}), \\
    {\mathbf{s}}_{pred} = StrokeReg({\mathbf{s}}_{init},I_{target}),
\end{align}
where $C_{init}$ signifies the blank canvas initialization and $I$ is the target image. In the following sections, we discuss each of the above modules in full detail.

\subsection{Sequential Planner}

\subsubsection{Reinforcement Learning Formulation}
\label{sec:rl_formulation}

The \emph{Sequential Planner} is modelled as a deep reinforcement learning agent which learns a painting policy $\pi$ predicting vectorized brushstroke parameters $\mathbf{a}_t$ from the current agent state $s_t$. The agent state $s_t$ at any timestep $t$ is modeled as the tuple $(C_t,I,t,\mathcal{G}_t,\mathcal{W}_t,\mathcal{S}_I,l)$, where $C_t$ is the canvas state, $I$ is the target image, $\mathcal{S}_I$ signifies the target-image saliency map, $l$ is the current painting layer (Sec.~\ref{sec:progressive_layering}), and $(\mathcal{G}_t,\mathcal{W}_t)$ represent the coarse and fine local attention windows for the painting agent respectively (Sec.~\ref{sec:seq_guidance}).

The canvas state $C_t$ is updated using a differentiable neural renderer module, which rasterizes the predicted brushstroke parameters $\mathbf{a}_t$ to output a brushstroke alpha map $S_{\alpha}(\mathbf{a}_t)$ and its colored rendering $S_{color}(\mathbf{a}_t)$. The canvas updates at each timestep $t$ are then computed as follows, 
\begin{align}
    C_{t+1} = C_t \odot (1-S_{\alpha}(\mathbf{a}_t)) + S_{color}(\mathbf{a}_t).
\end{align}

We next discuss further details regarding the above formulation which allows the painting agent to generate canvases while exhibiting a human-like painting process.

\subsubsection{Progressive Layering}
\label{sec:progressive_layering}
The human painting process is often progressive and multi-layered \cite{zhao2020painting,reyner2017layer}. That is, instead of painting everything on the canvas at once, humans often first paint a basic background layer before progressively adding each of the foreground objects on top of it (refer Fig.~\ref{fig:overview}). However, such a strategy is hard to learn using previous works which directly minimize the pixel wise distance the generated canvas $C_t$ and the target image $I$.

To this end, we propose a progressive layering strategy, which much like a human artist, allows the painted canvas to evolve in multiple successive layers. 
The objective of the painting agent in the first layer, is to paint a realistic background scene by trying to only focus on the non-salient (background) image areas. In doing so, the salient image regions are painted so as to maximize the efficiency of painting the background contents (\emph{e.g.} salient region corresponding to a bird sitting on a tree would be painted while focusing on tree leaves and branches, as in Fig.~\ref{fig:human_sim_qual}). Once the background layer is drawn, the painting agent in the successive layer then proceeds to add different foreground objects in a decreasing order of saliency. An illustration of a two layer painting process is shown in Fig.~\ref{fig:model_design}. The painting agent first draws a realistic background scene (by focusing only on background image contents like ground, grass, \emph{etc}.), before adding in the foreground objects in the second layer.

In order to achieve this layering process, we first divide the painting episode into multiple layers as follows,
\begin{align}
    C_{out} = \sum_{l=0}^{L-1} \  \sum_{t=1}^{T/L} C^l_t \odot (1-S_{\alpha}(\mathbf{a}^l_t)) + S_{color}(\mathbf{a}^l_t),
\end{align}
where $L=2$ is the number of layers\footnote{For simplicity, we primarily use $L=2$ in the main paper. Details on extending progressive layering to $L>2$ are provided in supp. material.}, $T$ is the episode length, $C_0^{l=0}$ signifies an empty initial canvas, and $C_0^{l=1}$ is initialized as the canvas output $C^{l=0}_{T/L}$ from the last layer.

Given canvas state $C_t$, input image $I$ and foreground saliency map $\mathcal{S}_I$, the layerwise painting style can then achieved be achieved by optimizing the following layered reward objective for each layer $l$, 
\begin{equation}
    \begin{aligned}
        r^{layer}_t (l) = \ & D(I \odot \mathcal{M}_I(l) , C_{t+1}\odot \mathcal{M}_I(l) ) \\
    - & D(I \odot \mathcal{M}_I(l),C_{t} \odot \mathcal{M}_I(l)),
    \end{aligned} \label{eq:layer_reward}
\end{equation}
where $D(I,C_t)$ is the joint conditional Wasserstein GAN \cite{arjovsky2017wasserstein} discriminator score for image $I$ and canvas $C_t$ \cite{huang2019learning}, and the layered-mask $\mathcal{M}_I(l)$ is defined as,
\begin{align}
    \mathcal{M}_I(l) = 1-\mathcal{S}_I \odot (1-l).
\end{align}

\subsubsection{Sequential Brushstroke Guidance}
\label{sec:seq_guidance}

Human painters often exhibit a localized spatial attention span while focusing on distinct image areas \cite{zhao2020painting}. This is in stark contrast with previous works which either compute stroke decomposition globally over the entire canvas or over a set of predefined grid regions \cite{huang2019learning,liu2021paint, zou2021stylized}. To this end, we propose a sequential brushstroke guidance strategy, which allows the reinforcement learning agent to shift its attention between different image regions through a sequence of coarse-to-fine attention windows $\{\mathcal{W}_0, \mathcal{W}_1 \dots \mathcal{W}_T\}$. The computation of the localized attention window $\mathcal{W}_t$ at any timestep $t$ is done in the following broad steps,

\textbf{Foreground object selection.}
The RL agent first selects the in-focus foreground object by predicting coordinates of a coarse global attention window $\mathcal{G}_t$. Given an input image $I$ with $N$ foreground objects, we model $\mathcal{G}_t = x^\mathcal{G}_t,y^\mathcal{G}_t,w^\mathcal{G}_t,h^\mathcal{G}_t$ as a convex combination of each of in-image object bounding box detections $\mathcal{B}_i \in \mathbb{R}^{4}, i \in [1,N]$.
    \begin{align}
        \mathcal{G}_t = \sum_{i=0}^N \alpha^t_i \ \mathcal{B}_i, \quad s.t. \ \  \forall t \ \  \sum_i \alpha^t_i = 1, \quad \alpha^t_i \geq 0.\label{eq:fg_obj_sel_2}
    \end{align}
where $\bm{\alpha}^t=\{\alpha^t_1, \  \dots \ \alpha^t_N\} \in \mathbb{R}^{N}$ are the spatial attention parameters predicted by the RL agent at timestep $t$. $\mathcal{B}_0$ represents an attention window over the entire canvas and is used while switching focus to background image areas.

\textbf{Local attention window selection.}
Within each object window $\mathcal{G}_t$, the agent further learns to sequentially shift its focus on different in-object features through a sequence of coarse-to-fine local attention windows $\mathcal{W}_t$. In particular, given the coarse object window coordinates $x^\mathcal{G}_t,y^\mathcal{G}_t,w^\mathcal{G}_t,h^\mathcal{G}_t$, the coordinates $\mathcal{W}_t = x^\mathcal{L}_t,y^\mathcal{L}_t,w^\mathcal{L}_t,h^\mathcal{L}_t$ for the finer localized attention windows are computed in a Markovian fashion as,
    \begin{align}
        x^\mathcal{L}_{t+1} = x^\mathcal{G}_{t+1} + (x^\mathcal{L}_{t} + \Delta x_t) \ w^\mathcal{G}_{t+1}, \\ y^\mathcal{L}_{t+1} = y^\mathcal{G}_{t+1} + (y^\mathcal{L}_{t} + \Delta y_t) \ h^\mathcal{G}_{t+1}, \\
        w^\mathcal{L}_{t+1} = (max(1-\Tilde{t},w_{min}) + \Delta w_t) \ w^\mathcal{G}_{t+1},\\ h^\mathcal{L}_{t+1} = (max(1-\Tilde{t},h_{min}) + \Delta h_t) \ h^\mathcal{G}_{t+1},
    \end{align}
where $\Tilde{t} \in [0,1]$ is the normalized episode timestep, $(w_{min},h_{min})$ are the minimum attention window dimensions and   $(\Delta \mathcal{W}_t = \Delta x_t,\Delta y_t,\Delta w_t,\Delta h_t)  \in \mathbb{R}^{4}$ are successive Markovian \cite{gagniuc2017markov} updates predicted by the RL agent.

\textbf{Brushstroke parameter adjustment}. Finally, the coordinates of local attention window are used to modify the predicted brushstroke parameters $\mathbf{a}^l_t$, so as constrain the painting agent to only draw within the localized attention window. 
This procedure can be expressed as,
\begin{align}
    \mathbf{a}^l_t \leftarrow ParamAdjustment (\mathbf{a}^l_t,\mathcal{W}_t).
\end{align}
Further details regarding the implementation of the adjustment function are provided in the supplementary material.

\subsubsection{Human-Consistency Penalties}
\label{sec:human_consistency}

Human artists inherently try to focus on spatially close image areas and try to avoid unnecessary spatial oscillations when painting a given image \cite{zhao2020painting}.  In this regard, while the Markovian adjustment procedure introduced in Sec.~\ref{sec:seq_guidance} ensures the spatial closeness of two consecutive local attention windows $\mathcal{W}_t$, unnecessary movements may still arise due to oscillations between different coarse attention windows $\mathcal{G}_t$. In order to prevent learning such stroke decompositions we introduce the following spatial penalty reward,
\begin{align}
    r^{spatial}_t = - \Vert \mathcal{G}_{t+1} - \mathcal{G}_{t} \Vert_F,
\end{align}
where $\Vert . \Vert_F$ represents the Frobenius norm. 

Similarly, human painting sequences are also characterized by the use of same / similar color patterns at consecutive timesteps \cite{zhao2020painting}. Thus, in order to mimic this behaviour we propose the following color transition penalty $r^{color}_t$,
\begin{align}
    r^{color}_t = - \Vert (R,G,B)_{t+1} - (R,G,B)_{t} \Vert_F,
\end{align}
where $(R,G,B)_{t}$ represents the brushstroke color prediction vector at timestep $t$.

\subsection{Brushstroke Regularization}
\label{sec:brush_reg}

\begin{figure*}[h!]
\begin{center}
\centerline{\includegraphics[width=0.88\linewidth]{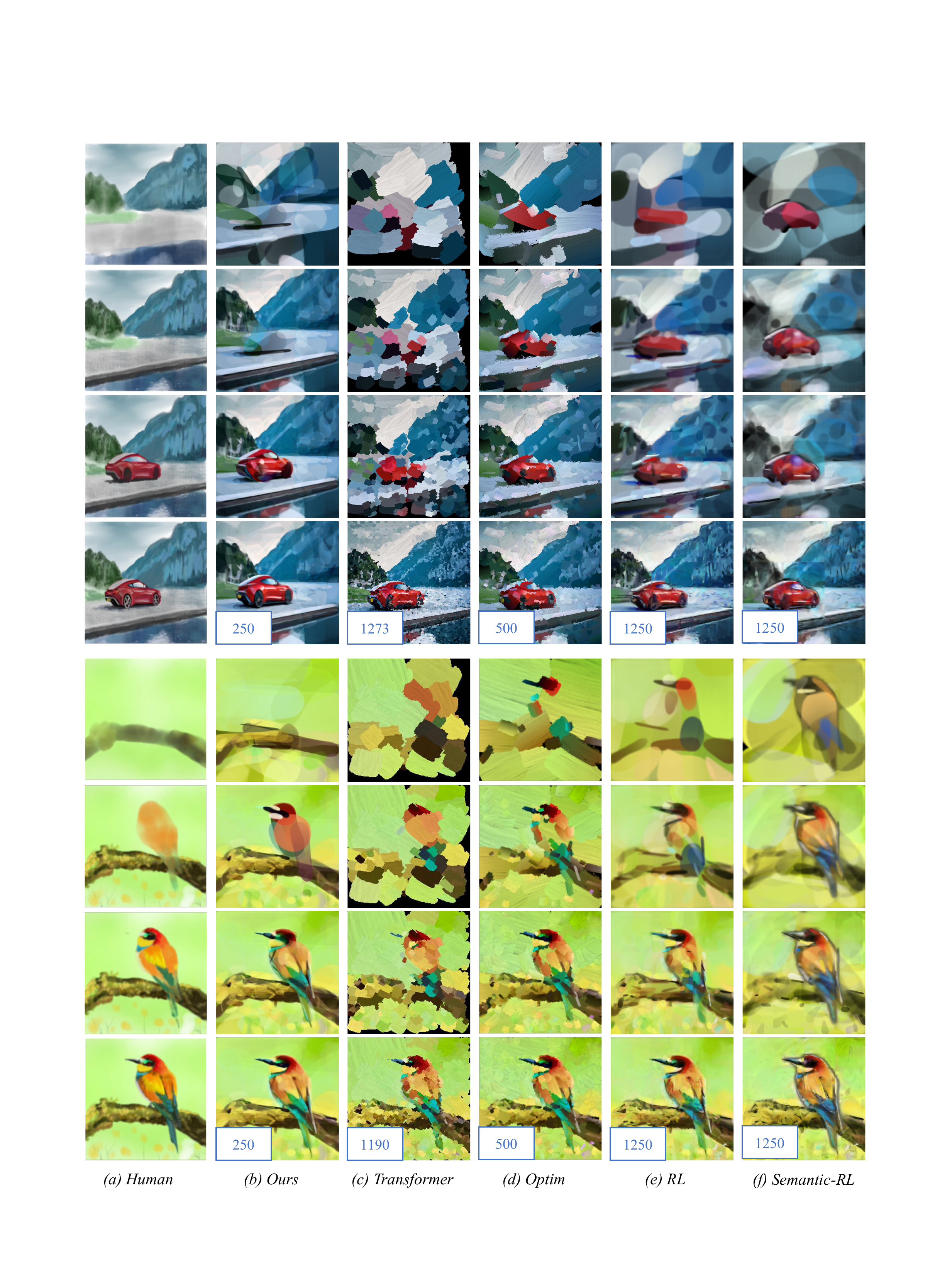}}
\caption{\emph{\textbf{Qualitative method comparison w.r.t resemblance with the human painting style.}} We compare different methods (b-f). All painting sequences are generated using a different brushstroke count (indicated in the boxes), so as to ensure similar pixel-wise reconstruction loss with the target image. The corresponding frames for each sequence are computed after $\sim$ 10\%, 40\%, 60\% and 100\% of the overall painting episode. We clearly see that our method
offers higher resemblance with the human painting style as compared to previous works.
}
\label{fig:human_sim_qual}
\end{center}
\vskip -0.3in
\end{figure*}

The current works on autonomous painting systems are often limited to using (an almost) fixed brush stroke budget irrespective of the complexity of the target image. Experiments reveal that this not reduces the efficiency of the generated painting sequence but also results in redundant / overlapping brushstroke patterns (refer Fig.~\ref{fig:human_sim_qual}) which impart an unnatural painting style to the final agent.

To address this, we propose an inference-time brushstroke regularization strategy which refines and removes redundancies from the initial brushstroke sequence predictions $\mathbf{s}_{init}$ to output the most efficient stroke decomposition $\mathbf{s}_{pred}$ for each test image. To do this, we first associate each brushstroke with the corresponding importance vector $\beta^l_t \in [0,1]$ by modifying the stroke rendering process as,
\begin{align*}
    C_{out} = \sum_{l=0}^{L-1} \  \sum_{t=1}^{T/L} C^l_t \odot (1-\beta^l_t \ S_{\alpha}(\mathbf{a}^l_t)) + \beta^l_t \ S_{color}(\mathbf{a}^l_t),
\end{align*}
where $\beta^l_t=Sign(x^l_t)$ and $x^l_t \sim \mathcal{N}(0,10^{-3})$ is randomly initialized from a normal distribution. 

We then use gradient descent in order to optimize the following loss function over both brushstroke parameters $\mathbf{a}^l_t$ and importance vectors $\beta^l_t$ (through $x^l_t$):
\begin{align}
    \mathcal{L}_{total}(\mathbf{a}^l_t,x^l_t) = \mathcal{L}_2 (I,C_{out}) + \gamma \ \sum_{l=0}^{L-1}\sum_{t=1}^{T/L} \Vert \beta^l_t \Vert_1, 
\end{align}
where the backpropagation gradients $\partial \beta^l_t/{\partial x^l_t}$ are computed as $\sigma (x^l_t) (1-\sigma(x^l_t))$, $\sigma(.)$ is the sigmoid function and $\gamma$ 
balances the weightage between brushstroke refinement and the need to use as few brushstrokes as possible. 

\begin{figure*}[h!]
\begin{center}
\centerline{\includegraphics[width=0.978\linewidth]{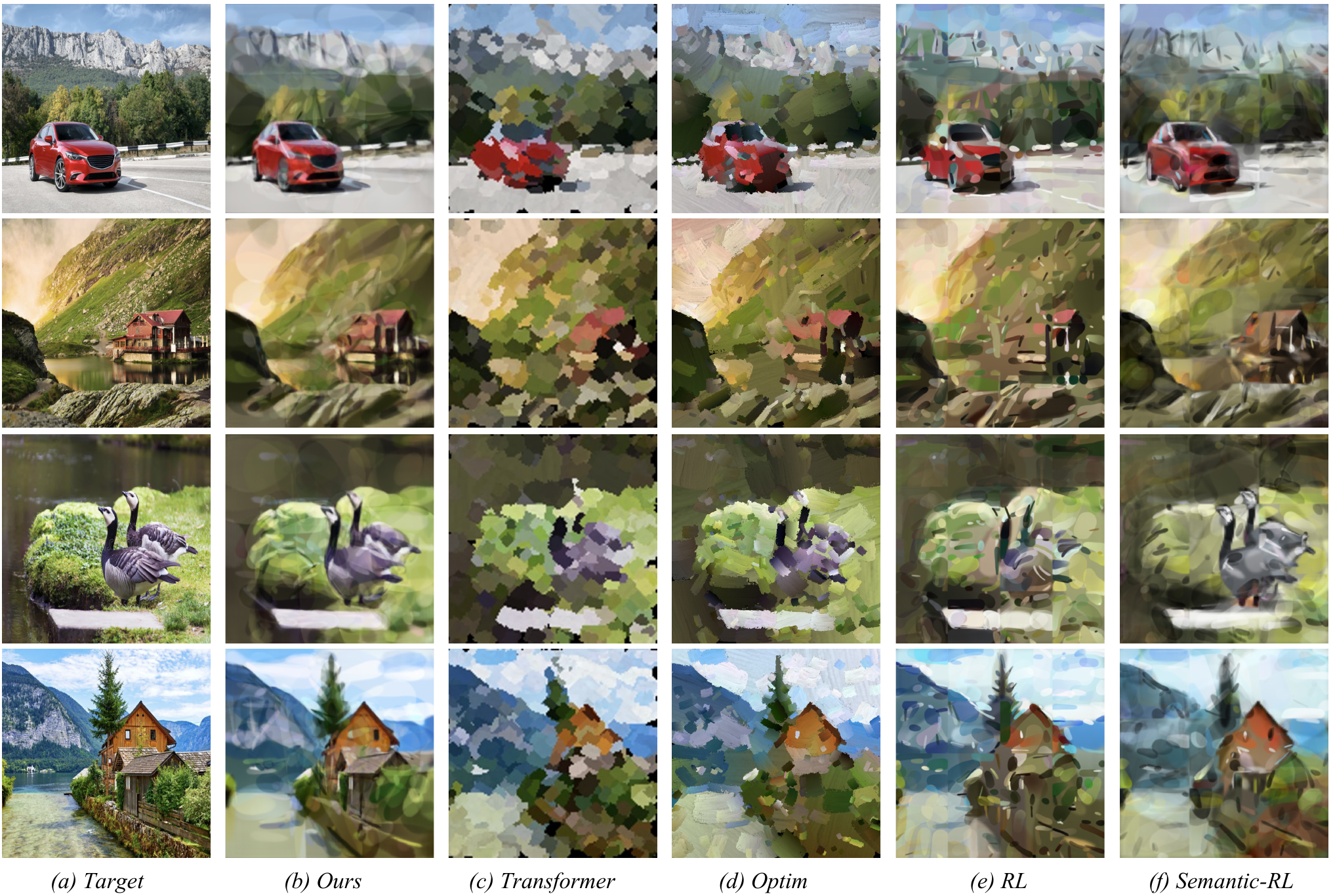}}
\vskip -0.05in
\caption{\emph{\textbf{Qualitative method comparison w.r.t painting efficiency.}} Comparing final canvas outputs while using $\sim$ 300 brushstrokes for (b) Ours, (c) Paint Transformer \cite{liu2021paint}, (d) Optim \cite{zou2021stylized}, (e) RL \cite{huang2019learning} and (f) Semantic-RL \cite{singh2021combining}. We observe that our approach results in more accurate depiction of the fine-grain features in the target image while using a low brushstroke count. Please zoom in for better comparison.
}
\label{fig:paint_eff_qual}
\end{center}
\vskip -0.35in
\end{figure*}

\section{Implementation Details}
\label{sec:training_details}

{\textbf{Neural renderer.}} In this paper, we primarily adopt the \emph{PixelShuffleNet} architecture from Huang \etal \cite{huang2019learning} while designing the neural differentiable renderer. While our approach is not limited to a particular rendering mechanism, we find that as opposed to the opaque brushstroke models used in \cite{zou2021stylized,liu2021paint}, the use of a more naturally blending brushstroke representation from  \cite{huang2019learning}, allows our method to mimic the human painting style in a more closer fashion.

{\textbf{Layered training.}} The use of progressive layering module requires conditionally training the painting agent policy at each layer while initializing the canvas state with the output from the last layer. In order to save computation time during training, we train the successive layer policies in consecutive batches while using the canvas output from the last layer. Furthermore, we only use $L=2$ layers at the training time. At inference time, the trained progressive layering policy can then be applied for $L>2$ layers by appropriately modifying the target image saliency maps. Please refer supplementary material for further details. 

{\textbf{Saliency and bounding box predictions}}. A key component of the Intelli-Paint pipeline is the sequential brushstroke guidance strategy which relies on the computation of object saliency and bounding box predictions.
In this work, we use a pretrained U-2-Net model \cite{u2net} model in order to compute foreground saliency predictions. The bounding box predictions are then computed as the union over bounding box outputs from pretrained Yolo-v5 \cite{yolo_v5} and the overall bounding box for the saliency prediction output.

{\textbf{Overall training.}} The RL-based \emph{Sequential Planner} agent is trained using the model-based DDPG algorithm \cite{huang2019learning} with the following overall reward function for each layer $l$,
\begin{align*}
    r_t^{overall}(l) = r^{layer}_t (l) + \mu \ r^{gbp}_t + \eta \ r^{spatial}_t + \gamma \  r^{color}_t,
\end{align*}
where $r^{gbp}_t$ is the guided backpropagation based focus reward from \cite{singh2021combining}. The hyperparameter $\mu$ is set to 10 for our experiments, while the human consistency penalty coefficients $\eta,\gamma$ are raised from $10^{-4}$ to $0.1$ in a linear schedule during the training process. The final RL agent is trained for a total of 5M iterations with a batch size of 128.

\section{Comparison with the state of the art}
\label{sec:results}

In this section, we provide extensive qualitative and quantitative results comparing our method with recent state of the art neural painting methods \cite{huang2019learning,zou2021stylized,singh2021combining,liu2021paint}. First, in Sec.~\ref{sec:painting_efficiency}, we demonstrate the improved painting efficiency of our method in generating detailed paintings when using limited number of brushstrokes. Second, we show that our method leads to painting sequences with increased resemblance with the human painting style (refer Sec.~\ref{sec:sim_human}). Finally, we provide a discussion of some assumptions that our method makes and analyse the robustness of our approach to potential violations of the same (refer Sec.~\ref{sec:limitations}).

\subsection{Painting Efficiency}
\label{sec:painting_efficiency}

\emph{\textbf{Qualitative Comparison.}} Fig.~\ref{fig:paint_eff_qual} shows a qualitative comparison between the generated canvases using a low budget of 300 brushstrokes per canvas. Note that due to grid-wise formulation for Paint Transformer \cite{liu2021paint} and Optim \cite{zou2021stylized}, the corresponding results are reported after $\sim$360 and 330 brushstrokes respectively. We clearly see that our method results in more accurate depictions of target image (\emph{e.g.} fine-grain features for car, hut, and birds in row 1-3 from Fig.~\ref{fig:paint_eff_qual}) when using limited number of brushstrokes.
In contrast, previous methods often lack an intelligent mechanism for efficient brushstroke distribution across the canvas which leads to poor performance when using a limited brushstroke budget. Interestingly, we also note that while Paint Transformer \cite{liu2021paint} exhibits improved inference speeds, it performs worse than previous methods like Optim \cite{zou2021stylized} when using a small number of brushstrokes.

\emph{\textbf{Quantitative Comparison.}} Table \ref{tab:paint_eff_quant} shows quantitative results on the quality of the finally generated canvases using $\sim$ 300 brushstrokes per canvas. The final results are reported in terms of both pixel wise $l_2$ distance $\mathcal{L}_{pixel}$ and perceptual similarity loss $\mathcal{L}_{pcpt}$ \cite{johnson2016perceptual} between the final canvas and the target image. The quantitative values show that our method helps in significantly lowering the distance metrics between the painted canvas and the target image as compared to previous works. In particular, we note that for CUB-Birds dataset \cite{WahCUB_200_2011}, our approach leads to a reduction of 47.1\%, 25.6\% 24.9\% and 38.2\% in the $\mathcal{L}_{pixel}$ distance metric as compared to RL\cite{huang2019learning}, Semantic-RL\cite{singh2021combining}, Optim \cite{zou2021stylized} and Paint Transformer \cite{liu2010single}, respectively.

\begin{table}
\begin{center}
\begin{tabular}{lcccc}
\toprule
\multirow{2}{*}{Method} & \multicolumn{2}{c}{Stanford Cars\cite{KrauseStarkDengFei-Fei_3DRR2013}} & \multicolumn{2}{c}{CUB-Birds\cite{WahCUB_200_2011}}\\
\cline{2-5}
 & $\mathcal{L}_{pixel}$ & $\mathcal{L}_{pcpt}$ & $\mathcal{L}_{pixel}$ & $\mathcal{L}_{pcpt}$\\
\hline
RL \cite{huang2019learning} & 97.85 & 0.67  & 96.14 & 0.76 \\
Semantic-RL \cite{singh2021combining} & 79.98 & 0.55 & 68.46 & 0.55\\
Optim \cite{zou2021stylized} & 76.52 & 0.54 & 67.90 & 0.53\\
Transformer \cite{liu2021paint} & 87.78 & 0.57 & 82.43 &  0.56 \\
\textbf{Ours} & \textbf{56.92} & \textbf{0.44} & \textbf{50.94} & \textbf{0.45}\\
\bottomrule
\end{tabular}
\end{center}
\vskip -0.2in
\caption{\emph{\textbf{Method comparison w.r.t painting efficiency.}} Our method results in much more accurate depictions of the target image while using a limited brushstroke budget.}
\label{tab:paint_eff_quant}
\vskip -0.1in
\end{table}

\subsection{Similarity with Human Painting Style}
\label{sec:sim_human}

\emph{\textbf{Qualitative Comparison.}} We demonstrate the practical applicability of our method to actual human users by qualitatively comparing the painting sequences generated by our method with those drawn by actual human artists. Results are provided in Fig.~\ref{fig:human_sim_qual}. We observe that our method bears high resemblance with the human painting style in terms of both layerwise painting evolution and localized attention. In contrast, previous state of the art methods often try to directly minimize the pixel wise distance between the painted canvas and the target image, thereby leading to intermediate canvas states which are less intelligible for a human user.

For instance, in the first example from Fig.~\ref{fig:human_sim_qual}, much like a human painter, our method first paints a realistic background representation (consisting of the sky, mountains, river and the ground) before drawing in the foreground car in a coarse-to-fine fashion. This results in a more human-like evolution of the painted canvas which can be easily relatable to actual human artists. In contrast, methods like Paint Transformer\cite{liu2021paint}, Optim \cite{zou2021stylized} and RL \cite{huang2019learning} directly make brushstrokes based on low-level image features (\emph{e.g.} red brushstrokes for the car in row-1 and head of the bird in row-5 from Fig.~\ref{fig:human_sim_qual}). This leads to more bottom-up painting sequences which are different from the human style.
Meanwhile, Semantic-RL \cite{singh2021combining} tries to paint both foreground and background regions in parallel, thereby lacking the semantic painting evolution exhibited by human users.

\emph{\textbf{Quantitative Comparison.}} We also report quantitative results demonstrating the human-likeliness of our approach as compared to previous works. To this end, we devise a human user study wherein each human participant is shown a series of paired painting sequences comparing our method with previous works. For each pair, the human subject is then asked to select the painting sequence which best resembles the human painting style. The user study was conducted across 50 unique Amazon Mechanical Turk subjects and a set of randomly chosen 100 painting sequences from the CelebA \cite{CelebAMask-HQ} and CUB-Birds \cite{WahCUB_200_2011} datasets. Results are reported in Table \ref{tab:human_sim_quant}. We clearly see that our method results in painting sequences which are perceived as a closer match for the human painting style by majority of human users.

\begin{table}
\begin{center}
\begin{tabular}{lc}
\toprule
 Method & Intelli-Paint Preference\\
\hline
RL \cite{huang2019learning} & 83.11 \% \\ 
Semantic-RL \cite{singh2021combining} & 69.09 \% \\ 
Optim \cite{zou2021stylized} & 75.41 \%\\
Transformer \cite{liu2021paint} & 86.50 \%\\
\bottomrule
\end{tabular}
\end{center}
\vskip -0.2in
\caption{\emph{\textbf{User-Study:}} Showing \% of painting samples for which human users prefer Intelli-Paint sequences over previous works.
In all the four studies, more users chose Intelli-Paint over the other.}
\label{tab:human_sim_quant}
\vskip -0.12in
\end{table}

\subsection{Robustness to Limitations / Assumptions}
\label{sec:limitations}

While our work results in significant improvements in both painting efficiency and human-likeliness of the generated painting sequence, it does build on some inherent assumptions. 
\textbf{\emph{First}}, we note that our method only mimics some \emph{general} aspects (progressive layering, coarse to fine localized attention) of the human painting process, and, thus does not claim to be calibrated to the fine-grain variations in the painting styles of each human artist. Nevertheless, as demonstrated in Table \ref{tab:human_sim_quant}, we find that our painting style is considered more relatable by majority of human users. 

\textbf{\emph{Second}}, our method relies on the computation of image saliency masks for allowing a human-like evolution of the painted canvas. Thus limitations of the pretrained U2-Net \cite{u2net} model become our limitations. Nevertheless, we note that failure to detect a particular salient object would simply lead to painting the corresponding region in the background layer, and thus does not affect the quality of the final canvas.

\textbf{\emph{Finally,}} to learn a human-like painting style, 
our method requires self-supervised training on a dataset of 
\emph{real} images. This is in contrast with Paint transformer \cite{liu2021paint} which performs self-training on an \emph{artificial} dataset, and Optim \cite{zou2021stylized} which does not require any training. That said, once trained we find that our method is able to generalize across a range of domains at inference time. For instance, we note that all results in Fig.~\ref{fig:paint_eff_qual}, \ref{fig:human_sim_qual} were generated using an Intelli-paint model trained only on the CUB-Birds \cite{WahCUB_200_2011} dataset.

\section{Conclusion}
\label{sec:conclusion}

In this paper, we emphasize that the practical merits of an autonomous painting system should be evaluated not only by the quality of generated canvas but also by the interpretability of the corresponding painting sequence by actual human artists. To this end, we propose a novel \emph{Intelli-Paint} pipeline which uses progressive layering to allow for a more human-like evolution of the painted canvas. The painting agent focuses on different image areas through a sequence of coarse-to-fine localized attention windows and is able to paint detailed scenes while using a limited number of brushstrokes. Experiments reveal that in comparison with previous state-of-the-art methods, our approach not only shows improved painting efficiency but also exhibits a painting style which is much more relatable to actual human users.

\newpage
{\small
\bibliographystyle{ieee_fullname}
\bibliography{egbib}
}

\appendix

\title{Supplementary Material \\ 
Intelli-Paint: Towards Developing Human-like Painting Agents}

\twocolumn[{
\maketitles
\begin{center}
    \vskip -0.2in
    \captionsetup{type=figure}
    \includegraphics[width=0.95\linewidth]{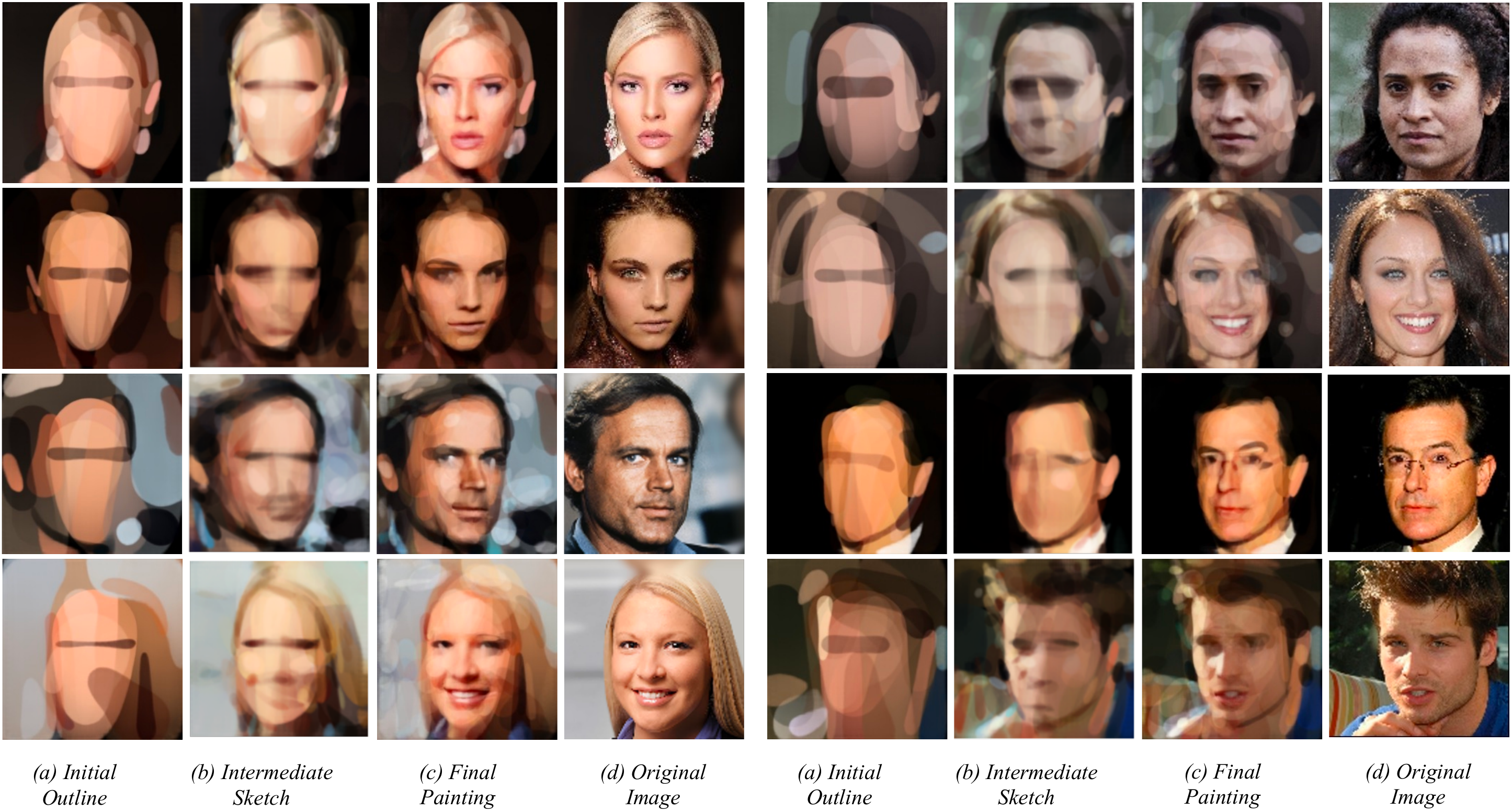}
    \captionof{figure}{\textbf{Artistic Painting Evolution.} In contrast with previous works 
    which directly try to minimize the $\mathcal{L}_{pixel}$ distance between the painted canvas and the original image (d), our method allows for a more human-like evolution of the painted canvas across different domains. For instance, for faces it first begins with a rough facial outline shown in (a). It then refines this outline further to paint the highly artistic human-like intermediate representation in (b), before finally adding all facial details (eyes, nose, lips) in the final painting (c).}
     \label{fig:artistic_evolution}
\end{center}
}]

\section{Analysis: Progressive Layering}
 The progressive layering module forms one of the key components of the \emph{Intelli-paint} pipeline. In this section, we further discuss different aspects of the progressive layering module. First, we demonstrate the human-like artistic painting evolution exhibited by our method for the facial domain in Appendix~\ref{sec:background_visualizations}. We then analyse the use of the progressive layering module as an effective alternative to performing foreground object removal for non-standard object masks (refer Appendix~\ref{sec:foreground_removal}). Finally, we provide a formulation for extending the proposed layering mechanism for $L>2$ layers (refer Appendix~\ref{sec:prog_layer_extension}).
 
\subsection{Artistic Painting Evolution}
\label{sec:background_visualizations}

By allowing the final painting agent to draw a given scene in multiple successive layers, the progressive layering strategy facilitates for a more human-like evolution of the painted canvas \cite{reyner2017layer}.
In this section, we analyse some intermediate canvas representations painted by our method, and compare its resemblance with the human-like ideation process for the facial domain.

\begin{figure*}[h!]
\vskip -0.1in
\begin{center}
\centerline{\includegraphics[width=0.93\linewidth]{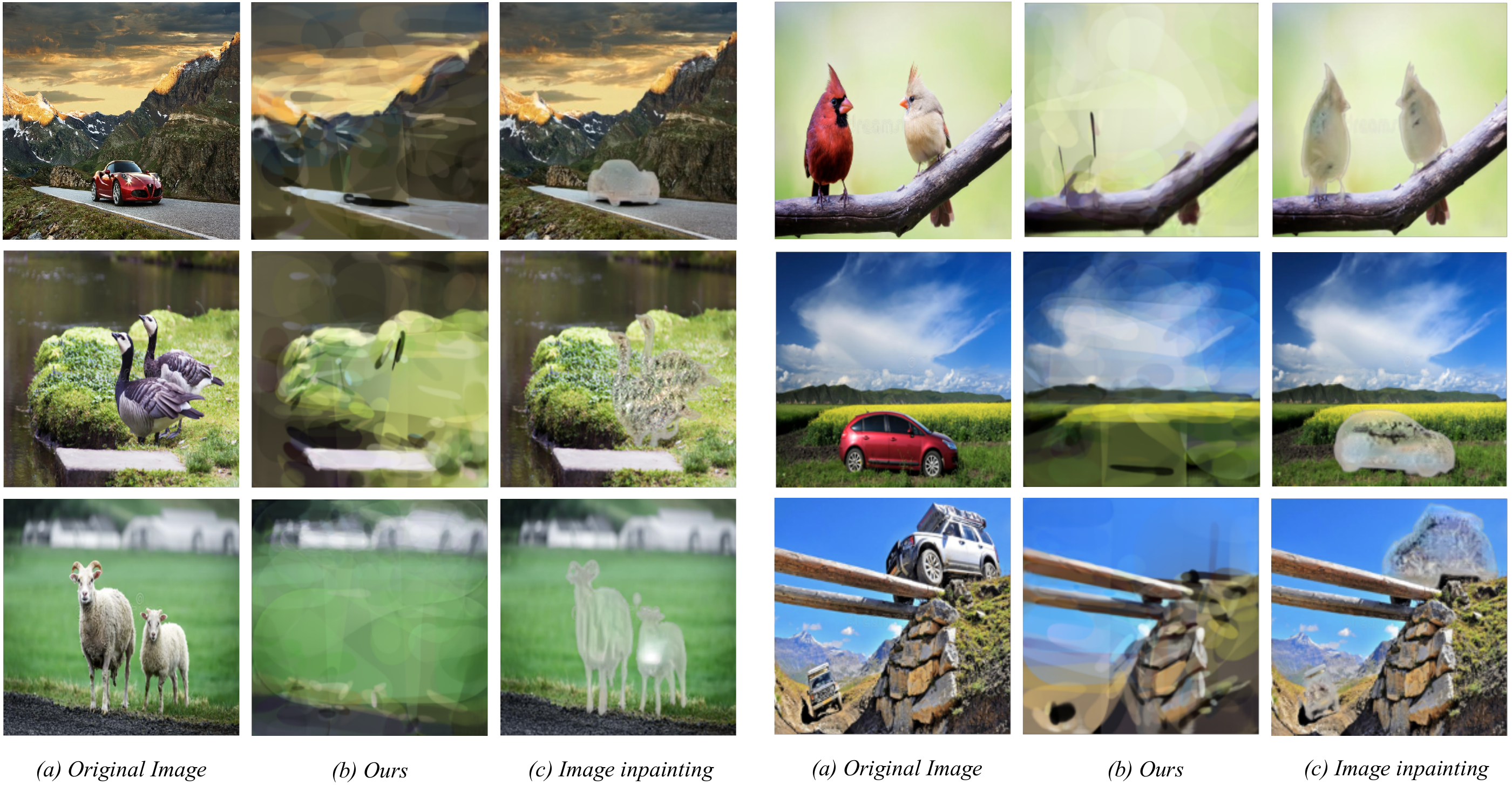}}
\vskip -0.05in
\caption{\textbf{Intelli-paint for foreground object removal.} Comparing foreground object removal using (b) progressive layering vs GAN-based image inpainting from (c) Yu \etal. \cite{yu2018generative}. Our method performs image inpainting using \emph{learning to paint} paradigm itself and results in much more natural background scene representations as opposed to GAN based image inpainting methods for non-standard object masks.}
\label{fig:fg_removal}
\end{center}
\vskip -0.4in
\end{figure*}

Results are shown in Fig.~\ref{fig:artistic_evolution}. We observe that instead of directly trying to minimize the pixel-wise distance between the painted canvas and the target image (as is done in previous works), our method takes a much more human-like approach to portrait generation. For instance, consider the second example (row-2) for the painting sequence in the right column. Instead of directly painting based on low-level features (\emph{e.g.} white brushstrokes for the mouth region), our method first draws a rough outline for the facial shape. It then refines this outline whilst indicating (but not drawing) the potential locations of important focalpoints (eyes, nose, lips) through appropriate facial shading. Only after the intermediate sketch has been set up, does it then proceed to add the fine-grain details on the exact facial features (eyes, nose, mouth, \emph{etc}). 

\subsection{Application: Foreground Removal}
\label{sec:foreground_removal}

The progressive layering module of the \emph{Intelli-Paint} pipeline also provides an effective alternative for performing foreground object removal using non-standard object masks. As compared to previous GAN based methods, our method performs image in-painting over the foreground regions using the \emph{learning to paint} paradigm itself. Results are shown in Fig.~\ref{fig:fg_removal}. We clearly see that the intermediate canvases generated using our method result in much more natural looking background scene representations. In contrast, GAN based image-inpainting methods show poor generalizability over large non-standard object masks and clearly reveal the presence of the foreground object in the final image predictions.

\subsection{Extending Progressive Layering}
\label{sec:prog_layer_extension}

The progressive layering formulation discussed in the main paper primarily divides the painting process into two broad layers (background and foreground). In this section, we show that the original two layered progressive formulation can be easily extended to $L>2$ layers by using ranked saliency maps \cite{siris_2020_ranked} for the target image $I$.

In particular, consider a ranked saliency map ${\mathcal{S}_I}$, such that $\{\mathcal{S}_I[k]\}_{k=1}^{L}$ indicate the salient regions for the target image $I$ ranked in increasing order of saliency.
The progressive layering strategy can then be trivially extended to $L>2$ layers by using the following layered-mask $\mathcal{M}_{I}(l)$ (refer Eq. 5, 6 in the main paper) for each layer $l$,
\begin{align}
    \mathcal{M}_{I}(l) = 1 - \bigcup_{k=1}^{L-l} \mathcal{S}_I[k].
\end{align}

\begin{figure*}[t]
\begin{center}
\centerline{\includegraphics[width=1\linewidth]{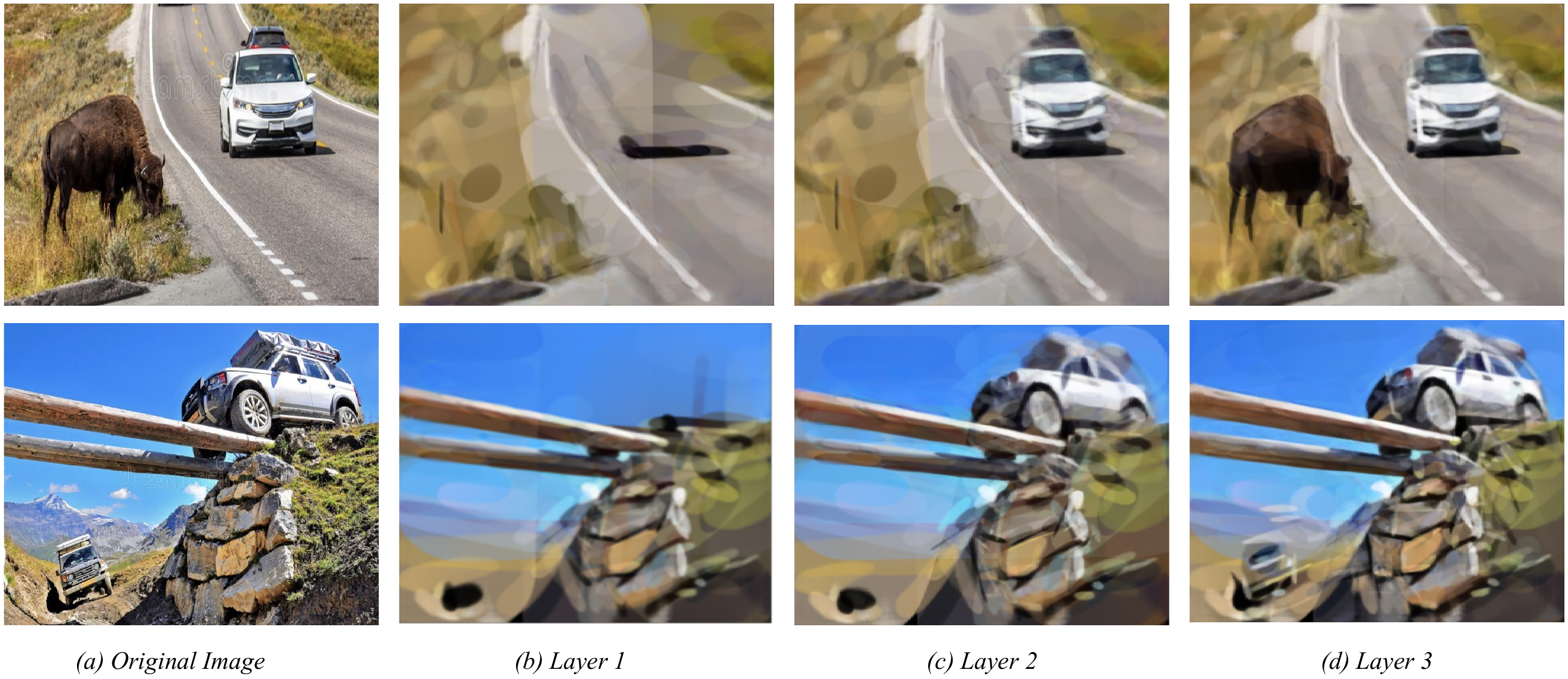}}
\caption{\textbf{Progressive Layering for $L>2$ layers.} The agent first begins by painting a realistic background scene. Once the background layer is drawn,  the painting agent in each successive layer then proceeds to add different foreground objects in decreasing order of saliency.}
\label{fig:prog_layering_mult}
\end{center}
\vskip -0.3in
\end{figure*}

Results for progressive painting sequences while using $L>2$ layers are shown in Fig.~\ref{fig:prog_layering_mult}. The painting process is divided in three layers. We observe that in the first layer, the painting agent proceeds to draw a realistic background scene on the canvas. In the second layer, it then proceeds to add the most salient object (white car) on to the canvas. Finally, in the third layer it shifts its attention to the least salient object (cow) in order to complete the final painting.

\section{Ablation Studies}
\label{sec:ablation studies}

The \emph{Intelli-paint} pipeline utilizes three main modules for mimicking the human painting style: 1) progressive layering, 2) sequential brushstroke guidance and 3) brushstroke regularization. In this section, we aim to understand the contribution of each of these modules in the development of a human-like painting process. However, since the main focus of our method is the recreation of the human artistic creation process, we argue that the effect of each module is best understood by visualizing the corresponding painting sequences. 
We thus refer the readers to the \href{https://1jsingh.github.io/intelli-paint}{official project webpage} for detailed visualizations on the importance of each module in the human-like painting process.
 
\section{Algorithm Details}
\label{sec:algorithm_details}
In this section, we further elaborate on some algorithm details which could not be included / fully explained due to space constraints in the main paper.

\subsection{Sequential Brushstroke Guidance}
\label{sec:seq_guidance}


The sequential brushstroke guidance strategy (discussed in Sec.~3.1.3 of main paper), allows the reinforcement learning based sequential planner agent to shift its attention between different image regions through a sequence of coarse-to-fine attention windows $\{\mathcal{W}_0, \mathcal{W}_1 \dots \mathcal{W}_T\}$. The computation of the localized attention window $\mathcal{W}_t$ at any timestep $t$ is done in the following broad steps.

\textbf{Foreground object selection.}
The RL agent first selects the in-focus foreground object by predicting coordinates of a coarse global attention window $\mathcal{G}_t$.  Given an input image $I$ with $N$ foreground objects, the computation of the coarse attention window $\mathcal{G}_t$ can be expressed through Algorithm \ref{alg:fg_select}.
\begin{algorithm}[h!]
	\caption{Foreground Object Selection}
    \textbf{Input}: Foreground selection convex coefficients $\bm{\alpha}^t$; In-image bounding box detections $\mathcal{B}_i \in \mathbb{R}^{4}, i \in [1,N]$.\\
    \textbf{Output}: Coarse object attention window $\mathcal{G}_{t}$.\\
    \textbf{Defaults}: Bounding box over the entire image $\mathcal{B}_0$.
    \begin{algorithmic}[1]
        \Function{ObjectSelect}{$\bm{\alpha}^t,\{\mathcal{B}_0,\dots \mathcal{B}_N\}$}
          \State $\{\alpha^t_0, \  \dots \ \alpha^t_N\} = \bm{\alpha}^t \in \mathbb{R}^{N+1}$;
          \State $\mathcal{G}_t = \sum_{i=0}^N \alpha^t_i \ \mathcal{B}_i$; 
          \State \Return $\mathcal{G}_{t}$.
        \EndFunction
	\end{algorithmic}
	\label{alg:fg_select}
\end{algorithm}

\textbf{Local attention window selection.}
Within each object window $\mathcal{G}_t$, the agent further learns to sequentially shift its focus on different in-object features through a sequence of coarse-to-fine local attention windows $\mathcal{W}_t$. In particular, given the coarse object window coordinates $x^\mathcal{G}_t,y^\mathcal{G}_t,w^\mathcal{G}_t,h^\mathcal{G}_t$, the coordinates $\mathcal{W}_t = x^\mathcal{L}_t,y^\mathcal{L}_t,w^\mathcal{L}_t,h^\mathcal{L}_t$ for the finer localized attention windows are computed using the Markov update function in Algorithm \ref{alg:markov_update}.

\begin{algorithm}[h!]
	\caption{Markov Updates for Local Attention Window}
    \textbf{Input}: Current coarse, local attention windows $(\mathcal{G}_{t},\mathcal{W}_{t})$; Markovian bounding box refinements $\Delta \mathcal{W}_{t}$.\\
    \textbf{Output}: Updated local attention window $\mathcal{W}_{t+1}$.\\
    \textbf{Defaults}: $w_{min}=h_{min}=0.2$.
    \begin{algorithmic}[1]
        \Function{MarkovUpdate}{$\mathcal{W}_t,\mathcal{G}_t,\Delta \mathcal{W}_t$}
        \\
          \State $x^\mathcal{G}_t,y^\mathcal{G}_t,w^\mathcal{G}_t,h^\mathcal{G}_t = \mathcal{G}_t$;
          \State $x^\mathcal{L}_t,y^\mathcal{L}_t,w^\mathcal{L}_t,h^\mathcal{L}_t = \mathcal{W}_t$;
          \State $\Delta x_t,\Delta y_t,\Delta w_t,\Delta h_t = \Delta \mathcal{W}_t$;
         \\
          \State $x^\mathcal{L}_{t+1} = x^\mathcal{G}_{t+1} + (x^\mathcal{L}_{t} + \Delta x_t) \ w^\mathcal{G}_{t+1}$;
          \State $y^\mathcal{L}_{t+1} = y^\mathcal{G}_{t+1} + (y^\mathcal{L}_{t} + \Delta y_t) \ h^\mathcal{G}_{t+1}$;
          \State $w^\mathcal{L}_{t+1} = (max(1-\Tilde{t},w_{min}) + \Delta w_t) \ w^\mathcal{G}_{t+1}$;
          \State $h^\mathcal{L}_{t+1} = (max(1-\Tilde{t},h_{min}) + \Delta h_t) \ h^\mathcal{G}_{t+1}$;
          \\
          \State $\mathcal{W}_{t+1} = x^\mathcal{L}_{t+1},y^\mathcal{L}_{t+1},w^\mathcal{L}_{t+1},h^\mathcal{L}_{t+1}$;
          \State \Return $\mathcal{W}_{t+1}$.
          \\
        \EndFunction
	\end{algorithmic}
	\label{alg:markov_update}
\end{algorithm}

\textbf{Brushstroke parameter adjustment}. Finally, the coordinates of local attention window are used to modify the predicted brushstroke parameters $\mathbf{a}^l_t$, so as constrain the painting agent to only draw within the localized attention window. 
This procedure can be expressed as,
\begin{align}
    \mathbf{a}^l_t \leftarrow \textsc{ParamAdjust} (\mathbf{a}^l_t,\mathcal{W}_t).
\end{align}

Assuming that $\mathbf{a}^l_t$ at each timestep $t$, depicts the parameters of a quadratic Bézier curve \cite{huang2019learning,singh2021combining} as follows, 
\begin{align}
    \mathbf{a}^l_t = (x_0,y_0,x_1,y_1,x_2,y_2,z_0,z_2,w_0,w_2,r,g,b),
\end{align}
where the first 10 parameters depict stroke position, shape and transparency, while the last 3 parameters form the \emph{rgb} representation for the stroke color. The parameter adjustment function is then implemented as per Algorithm \ref{alg:param_adjust}.

\begin{algorithm}[h!]
	\caption{Parameter Adjustment Function}
    \textbf{Input}: Initial vectorized brushstroke prediction $\mathbf{a}^l_t$; current local attention window coordinates $\mathcal{W}_t$.\\
    \textbf{Output}: Modified brushstroke prediction vector $\mathbf{a}^l_t$.
    \begin{algorithmic}[1]
        \Function{ParamAdjust}{$\mathbf{a}^l_t,\mathcal{W}_t$}
        \\
          \State $x_0,y_0,x_1,y_1,x_2,y_2,z_0,z_2,w_0,w_2,r,g,b = \mathbf{a}^l_t$;
          \State $x^\mathcal{L}_t,y^\mathcal{L}_t,w^\mathcal{L}_t,h^\mathcal{L}_t = \mathcal{W}_t$; 
        \\
        \State $x_0 = x^\mathcal{L}_t + x_0 \cdot w^\mathcal{L}_t$;
        \State $y_0 = y^\mathcal{L}_t + y_0 \cdot h^\mathcal{L}_t$;
        \State $x_2 = x^\mathcal{L}_t + x_2 \cdot w^\mathcal{L}_t$;
        \State $y_2 = x^\mathcal{L}_t + y_2 \cdot h^\mathcal{L}_t$;
        \State $w_0 = w_0 \cdot avg(w^\mathcal{L}_t,h^\mathcal{L}_t)$;
        \State $w_2 = w_2 \cdot avg(w^\mathcal{L}_t,h^\mathcal{L}_t)$;
        \\
          \State $\mathbf{a}^l_t = x_0,y_0,x_1,y_1,x_2,y_2,z_0,z_2,w_0,w_2,r,g,b$;
          \State \Return $\mathbf{a}^l_t$.
        \\
        \EndFunction
	\end{algorithmic}
	\label{alg:param_adjust}
\end{algorithm}

\subsection{Brushstroke Regularization}

The current works on autonomous painting systems are often limited to using (an almost) fixed brush stroke budget irrespective of the complexity of the target image. Experiments reveal that this not only reduces the efficiency of the generated painting sequence but also results in redundant / overlapping brushstroke patterns (refer main paper) which impart an unnatural painting style to the final agent.

To address this, we propose an inference-time brushstroke regularization strategy which refines and removes redundancies from the initial brushstroke sequence predictions $\mathbf{s}_{init}$ to output the most efficient stroke decomposition $\mathbf{s}_{pred}$ for each test image. This regularization procedure can be summarized through Algorithm \ref{alg:stroke_reg}.

\begin{algorithm}[h!]
	\caption{Brushstroke Regularization Function}
    \textbf{Input}: A target image $I$; initial brushstroke sequence $\mathbf{s}_{init} = \{\mathbf{a}^l_t \mid 0\leq l \leq L-1, 0\leq t \leq T/L\}$.\\
    \textbf{Output}: Refined brushstroke sequence $\mathbf{s}_{pred}$.\\
    \textbf{Defaults}: Number of layers $L$; episode length $T$; number of iterations $M$, $C_{init}=C_0^{l=0}=\textsc{BlankCanvas}$;
    \begin{algorithmic}[1]
        \Function{StrokeReg}{$\mathbf{s}_{init}, I$}:
          \State $\{\mathbf{a}^l_t \mid 0\leq l \leq L-1, 0\leq t \leq T/L\} = \mathbf{s}_{init}$;
          \State $x^l_t \sim \mathcal{N}(0,10^{-3}) \quad \forall l, \forall t$;
          \For{$0 \leq i \leq M$}
          \State $\beta^l_t=\textsc{Sign}(x^l_t) \quad \forall t,l$;
          \State\vspace*{-\baselineskip}
            \begin{fleqn}[\dimexpr\leftmargini+10pt]
            \setlength\belowdisplayskip{0pt}
            \begin{equation*}
                \begin{multlined}[c]
                  C_{out} = \sum_{l=0}^{L-1} \  \sum_{t=1}^{T/L} C^l_t \odot (1-\beta^l_t \ S_{\alpha}(\mathbf{a}^l_t)) \\ + \  \beta^l_t \ S_{color}(\mathbf{a}^l_t);
                \end{multlined}
            \end{equation*}
            \end{fleqn} \vspace{0.1in}%
            \State $\mathcal{L}_{total} = \mathcal{L}_2 (I,C_{out}) + \gamma \ \sum_{l=0}^{L-1}\sum_{t=1}^{T/L} \Vert \beta^l_t \Vert_1$;
            \State\vspace*{-\baselineskip}
            \begin{fleqn}[\dimexpr\leftmargini+10pt]
            \setlength\belowdisplayskip{0pt}
            \begin{equation*}
                  a^l_t \leftarrow a^l_t - \frac{\partial \mathcal{L}_{total}}{\partial a^l_t}\quad \forall t,l;
            \end{equation*}
            \end{fleqn}
            \State\vspace*{-\baselineskip}
            \begin{fleqn}[\dimexpr\leftmargini+10pt]
            \setlength\belowdisplayskip{0pt}
            \begin{equation*}
                  x^l_t \leftarrow x^l_t - \frac{\partial \mathcal{L}_{total}}{\partial x^l_t}\quad \forall t,l;
            \end{equation*}
            \end{fleqn}
          \EndFor
          \State $\mathbf{s}_{pred} = \{ \beta^l_t \cdot \mathbf{a}^l_t \mid 0\leq l \leq L-1, 0\leq t \leq T/L\} $;
          \State \Return $\mathbf{s}_{pred} $.
        \EndFunction
	\end{algorithmic}
	\label{alg:stroke_reg}
\end{algorithm}

\subsection{Overall Inference Algorithm}
\label{sec:inference_algorithm}

The overall inference algorithm for the \emph{Intelli-paint} pipeline can now be summarized as per Algorithm \ref{alg:inference}.
\begin{algorithm}[h!]
	\caption{Inference Algorithm for Intelli-Paint}
    \textbf{Input}: A target image $I$; image-saliency map $\mathcal{S}_I$; number of layers $L$; painting episode length $T$.\\
    \textbf{Required}: RL-based sequential planner $\textsc{POLICY}$; 
    \begin{algorithmic}[1]
        \State $\mathcal{W}_{0}= x^\mathcal{L}_0,y^\mathcal{L}_0,w^\mathcal{L}_0,h^\mathcal{L}_0 = (0,0,1,1)$;
        \State $C_{init}=C_0^{l=0}=\textsc{BlankCanvas}$;
        \For{$0\leq l \leq L-1$}
            \For{$0\leq t \leq T/L$}
                \State $s_t = (I,C^l_t,\mathcal{G}_t,\mathcal{W}_t,\mathcal{S}_I,l)$;
                \State $\mathbf{a}^l_t, \bm{\alpha^t}, \Delta \mathcal{W}_t = \textsc{Policy}(s_t)$;
                \State $\mathcal{G}_t = \textsc{ObjectSelect}(\bm{\alpha}^t,\{\mathcal{B}_0,\dots \mathcal{B}_N\})$;
                \State $\mathcal{W}_t = \textsc{MarkovUpdate}(\mathcal{G}_t,\mathcal{W}_{t-1},\Delta \mathcal{W}_t)$;
                \State $\mathbf{a}^l_t \leftarrow \textsc{ParamAdjust} (\mathbf{a}^l_t,\mathcal{W}_t)$;
                \State $C^l_{t+1} = C^l_t \odot (1-S_{\alpha}(\mathbf{a}^l_t)) + S_{color}(\mathbf{a}^l_t)$;
            \EndFor
        \EndFor
        \State $\mathbf{s}_{init} = \{\mathbf{a}^l_t \mid 0\leq l \leq L-1, 0\leq t \leq T/L\}$;
        \State  ${\mathbf{s}}_{pred} = \textsc{StrokeReg}({\mathbf{s}}_{init},I_{target})$;\\
        \Return $\mathbf{s}_{pred}$.
	\end{algorithmic}
	\label{alg:inference}
\end{algorithm}

\newpage
\section{Human User Study}
\label{sec:user_study_details}

\begin{figure}[h!]
\vskip -0.1in
\begin{center}
\centerline{\includegraphics[width=1\linewidth]{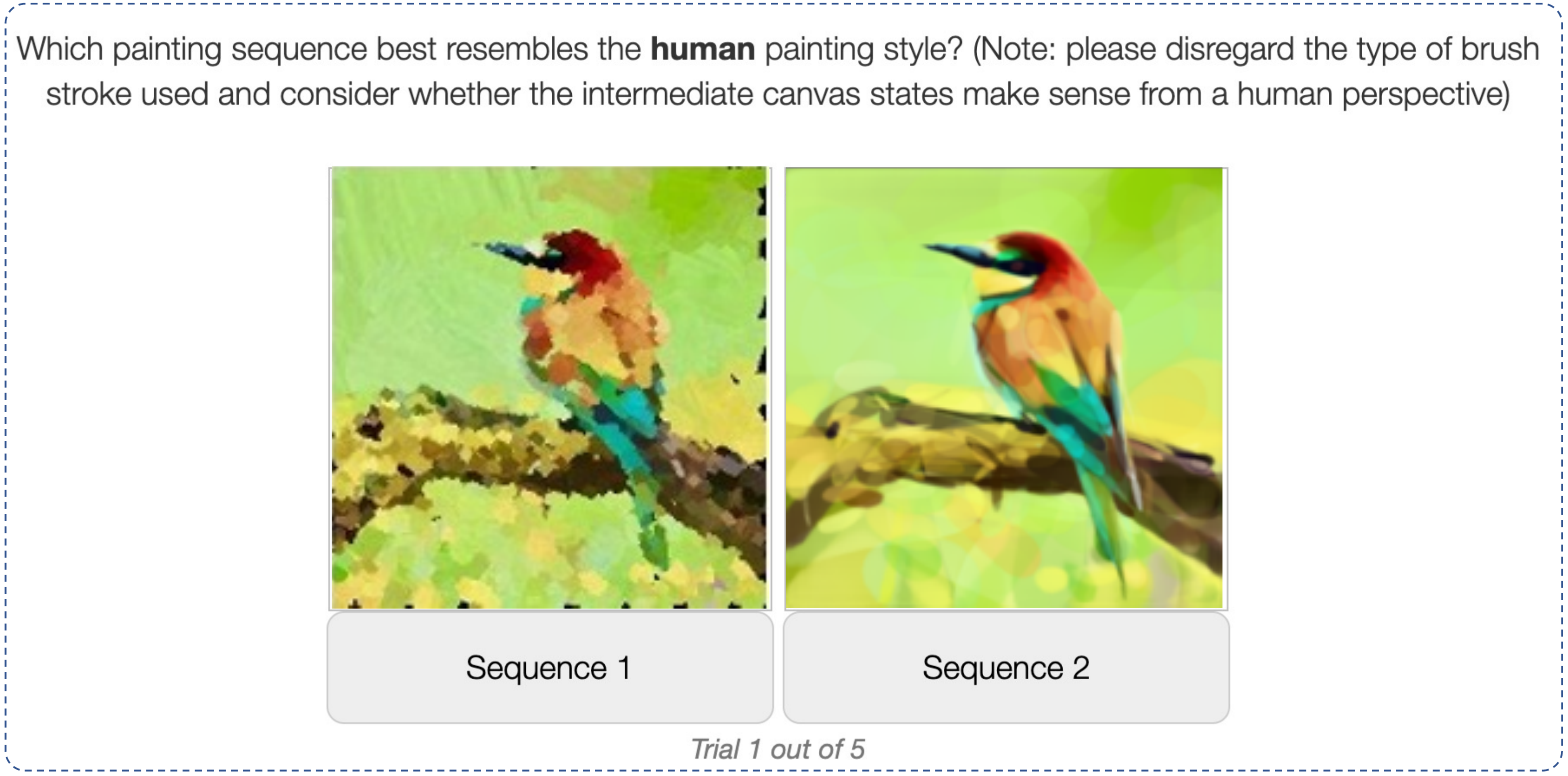}}
\caption{\textbf{Human User Study Interface.} Each user is shown two painting sequences (\emph{gif} format) and asked to select the one which would be most interpretable by actual human artists.}
\label{fig:user_study}
\end{center}
\vskip -0.35in
\end{figure}

In this section, we provide details about the human data collection process which is used to quantitatively demonstrate the human-likeliness of our approach as compared to previous state of the art \cite{huang2019learning,singh2021combining,zou2021stylized,liu2021paint}.

{\textbf{Basic Setup.}}
The user-study was conducted across 50 unique Amazon Mechanical Turk \cite{crowston2012amazon} subjects wherein each human participant is shown a series of paired painting sequences comparing our method with previous works. For each pair, the human subject is then asked to select the painting sequence which best resembles the human painting style. Each painting sequence is presented as a \emph{gif} image with a total duration of 10 seconds. Fig.~\ref{fig:user_study} illustrates the basic interface setup for the user-study.

{\textbf{Filtering process.}} Since the choice on human-likeliness of a painting sequence is subjective, we found that unfiltered collection of data from Amazon M-turk can lead to quite noisy responses, wherein many users simply select the responses at random in order to quickly collect the assignment reward. To address this, we take the following measures for avoiding data collection noise. First, we limit the data collection to human subjects with a HIT rate \cite{crowston2012amazon} greater than 90\%. In order to further refine the quality of collected data, we also limit the responses to users having a bachelors degree or equivalent. Furthermore, we intentionally use a repeated seed comparison for each sequence of paired comparisons shown to a particular user. Responses of users who respond differently to this repeated comparison are discarded while reporting the final results in the main paper.

\section{Inference Time Analysis}
\label{sec:inference_time_analysis}

While the ability to closely mimic the human artistic creation process determines the practical merits of a stroke based rendering approach \cite{huang2019learning,zou2021stylized,liu2021paint} over pixel-based image generation methods, the time taken to generate the same can help us better understand the practical usability of a method. In this section, we provide an analysis comparing the inference time required for our method with previous works \cite{huang2019learning,zou2021stylized,singh2021combining,liu2021paint}. In order to facilitate a better understanding for inference-time requirements of our approach we define two different variations of our method.

{\textbf{Ours (with StrokeReg)}}. This indicates the default Intelli-paint pipeline consisting of a \emph{Sequential Planner} for human-like painting sequence initialization and \emph{Stroke Regularizer} for predicting the most efficient brushstroke decomposition for each test image $I$. 

{\textbf{Ours (w/o StrokeReg)}}. We note that since the stroke regularization process is based on performing gradient descent across the entire painting trajectory, it is considerably slower than the \emph{Sequential Planner} agent. Also, since performing brushstroke compression may not be necessary for each potential application, we define a variant of \emph{Intelli-paint} which only relies on sequential predictions from the \emph{Sequential Planner} to learn a human-like painting process.

Results comparing inference time between different methods are shown in Table \ref{tab:inference_time}. All values are computed using the official implementations from respective authors \cite{huang2019learning,singh2021combining,zou2021stylized,liu2021paint} on a single Nvidia V100 GPU. We observe that since our method requires significantly fewer brushstrokes than Paint Transformer \cite{liu2021paint}, RL\cite{huang2019learning} and Semantic-RL \cite{singh2021combining}, it performs faster than the above when not using stroke regularization. Similarly, while both \emph{Ours (with StrokeReg)} and Optim \cite{zou2021stylized} require gradient descent based optimization, our method is considerably faster since it relies on high-quality initializations from the \emph{sequential planner} agent. In contrast, Optim \cite{zou2021stylized} typically begins with a random or heuristic-based brushstroke initialization which takes the gradient descent optimization longer to converge.

\begin{table}
\begin{center}
\begin{tabular}{lc}
\toprule
Method & Inference Time (\emph{\textbf{s}})\\
\hline
RL \cite{huang2019learning} & 2.317 s \\ 
Semantic-RL \cite{singh2021combining} & 2.631 s \\ 
Optim \cite{zou2021stylized} & 416.7 s\\
Transformer \cite{liu2021paint} & 1.154 s\\
Ours (\emph{with StrokeReg}) & 72.21 s \\
Ours (\emph{w/o StrokeReg}) & \textbf{0.948} s\\
\bottomrule
\end{tabular}
\end{center}
\vskip -0.2in
\caption{\textbf{Method comparison w.r.t. inference time.} We observe that since our method requires significantly fewer brushstrokes than Paint Transformer \cite{liu2021paint}, RL\cite{huang2019learning} and Semantic-RL \cite{singh2021combining}, it performs faster than the above when not using stroke regularization. Similarly, while both \emph{Ours (with StrokeReg)} and Optim \cite{zou2021stylized} require gradient descent based optimization, our method is considerably faster since it relies on high-quality initializations from the \emph{sequential planner} agent. In contrast, Optim \cite{zou2021stylized} typically begins with a random or heuristic-based brushstroke initialization which takes the gradient descent optimization longer to converge.}
\label{tab:inference_time}
\vskip -0.12in
\end{table}

\end{document}